%% file: main.tex
\documentclass[a4paper,fleqn]{cas-dc}

\usepackage{xcolor}
\input{acronyms}
\usepackage{graphicx}
\usepackage[font=footnotesize]{caption}
\usepackage{subcaption}
\usepackage{dsfont}

\definecolor{blue}{rgb}{0.21,0.49,0.74}

\usepackage[capitalize]{cleveref}
\crefname{section}{Sec.}{Secs.}
\Crefname{section}{Section}{Sections}
\Crefname{table}{Table}{Tables}
\crefname{table}{Tab.}{Tabs.}

\begin{document}

\title [mode = title]{Enhancing Vehicle Detection under Adverse Weather Conditions with Contrastive Learning}                      


\author[1]{Boying Li}
\ead{boying.li@ltu.se}

\author[1]{Chang Liu}
\ead{chang.liu@ltu.se}

\author[1]{Petter Ky\"{o}sti}
\ead{petter.kyosti@ltu.se}

\author[1]{Mattias \"{O}hman}
\ead{mattias.ohman@ltu.se}

\author[1]{Devashish Singha Roy}
\ead{devashish.singharoy@ltu.se}

\author[1]{Sofia Plazzi}
\ead{sofia.plazzi@ltu.se}

\author[2]{Olle Hagner}
\ead{olle.hagner@smartplanes.se}

\author[1]{Hamam Mokayed}
\ead{hamam.mokayed@ltu.se}

\affiliation[1]{organization={Electrical and Space Engineering},
    institution={Lule\aa\ University of Technology},
    city={Lule\aa},
    country={Sweden}
}
\affiliation[2]{organization={SmartPlanes AB, Skellefte\aa\, Sweden},
    country={Sweden}
}
\cortext[1]{Corresponding author}


\maketitle


\input{abstract_chang_v3}

\input{intro_chang_v2}

\input{relatedwork}

\input{method_v2_chang}

\input{experiments_251223}

\input{limitation}

\input{conclusion}

\section*{Acknowledgment}
This work was supported by the MARTINA project, that funded by the European Regional Development Fund, Region Norrbotten, Skellefte\aa\  Municipality, and Lule\aa\ Municipality.

\bibliographystyle{IEEEtran}
\bibliography{references}

\end{document}

%% file: acronyms.tex
\newcommand{\sideloadframework}{SCLA}
\newcommand{\SideloadFrameworkFull}{Sideload-Contrastive-Learning-Adaption}

%% file: abstract_chang_v3.tex
\begin{abstract}
Aside from common challenges in aerial object detection, i.e, small, sparse targets and computational power limitations, detecting vehicles from UAV images in the Nordic regions faces strong visibility challenges and domain shifts caused by diverse levels of snow coverage. Although data annotation is expensive, unannotated data is cheaper to collect by simply flying the drones. Hence, we propose a \SideloadFrameworkFull{}~(\sideloadframework{}) framework to improve lightweight detection model performance using unannotated data. 
We first pretrain a CNN-based representation extractor through contrastive learning on the unannotated data. Specifically, to address the unique challenges from the large image size and small objects, we propose the Feature-Map-Patch Contrastive Learning (FM-PaCL) method, which trains feature representation on patch level instead of global image level. Then, we fuse features from the FM-PaCL backbone and a frozen YOLO11n backbone in the fine-tuning stage for the detection task, to also utilize upstream representation learned from the COCO dataset.
Our proposed \sideloadframework{} framework improves the detection performance by 8.9\% in terms of mAP50 on the NVD dataset against the Yolo11n baseline. Code will be available upon acceptance. 

\end{abstract}

%% file: intro_chang_v2.tex
\section{Introduction}
\label{sec:intro}
Vehicle detection using UAVs ~(Unmanned Aerial Vehicle) offers greater coverage and flexibility than ground-based systems, but also presents unique challenges, such as overhead perspectives, small object scales, and weak visual features.
These challenges are amplified under snowy conditions, where visibility is reduced and vehicles covered by snow tend to blend with the background. In addition to common domain gaps encountered in object detection tasks, such as those caused by illumination changes and geographic variation, differences in snow coverage over vehicles and background objects introduce additional domain gaps. 

The Nordic Vehicle Dataset (NVD)~\cite{nvd}, collected in Northern Sweden, showcases such challenges. The NVD dataset includes both annotated data and unannotated data under different weather conditions and varsious snow coverage levels. 
While several studies have already utilized the annotated part of the NVD datasets~\cite{mokayed2024fractional, nvddetr, mokayed2024enhancing}, yet there has been no research utilizing the unannotated part.
Given that data annotation is time-consuming and costly, especially for video object detection which requires dense labeling of bounding boxes for objects in each frame~\cite{lin2014microsoft}, annotation in the NVD dataset is scarce. With limited annotated data, existing small models like Fast R-CNN~\cite{ren2016faster} and YOLO family~\cite{jiang2022review} are more susceptible to domain gaps and often suffer from poor generalization performance~\cite{mokayed2024enhancing}.  
To overcome the annotation scarcity and further enhance vehicle detection performance without additional annotation costs, we therefore propose a \SideloadFrameworkFull{}~(\sideloadframework{})  framework, which can significantly enhance the YOLO11n~\cite{yolo11} performance through contrastive learning and adaptive feature fusion. 

First, we propose the Feature-Map-Patch Contrastive Learning (FM-PaCL) method, which trains a CNN~(Convolutional Neural Network) feature extractor on the unannotated data with contrastive learning. 
This approach allows us to utilize a diverse range of unannotated data to build domain-robust feature extractors. 
Unlike existing contrastive learning methods that operate on global image views~\cite{chen2020simple,he2020momentum} or image patches~\cite{mukhoti2023open, yun2022patch, boserup2022efficient} , FM-PaCL enables the CNN feature extractor learn from the structured feature-map grid.
By contrasting spatially aligned patches across photometric augmented views, it preserves fine-grained local information critical for detecting small objects. 
To the best of our knowledge, no existing framework performs contrastive learning on overlapping patches extracted from intermediate CNN feature maps tailored for small-object representation. 

However, directly applying contrastive learning to the backbone, as a pretraining stage, leads to forgetting~\cite{forget} of the feature descriptors learned on the COCO dataset~\cite{lin2014microsoft}, which leads to a noticeable drop in performance compared to simply freezing YOLO11n's backbone (See Tab.~\ref{tab:ablativeframe}). This formulates the second key challenge of integrating the domain specific features descriptors learned from the unannotated data with the domain agnostic feature descriptors learned from the upstream COCO dataset. 
Since the representations learned through contrastive learning may not be naturally aligned with those from YOLO11n’s coco pretraining, traditional static fusion methods are also proven ineffective (See Fig.~\ref{fig:results}).

To tackle this, we propose a dynamic fusion approach, where the model learns to combine features during training, allowing it to adaptively assign importance to each channel (indivdual feature) based on input data and task objectives.

Our proposed \sideloadframework{} framework, as a system, yields a 8.9\% gain on the NVD dataset using the upstream protocol~\cite{nvd}, and a 2.8\% gain on a more practical split where training, testing, and validation are conducted on completely different videos. 






Our main contributions are summarized as follows:
\begin{enumerate} 
\item We propose a \SideloadFrameworkFull{} (\sideloadframework{}) framework, which allows us to use unannotated videos
to boost the Nordic vehicle detection performance under adverse weather
by 8.9\% in mAP50.

\item We propose a patch level contrastive learning framework, directly utilizing patches on featuremap level rather than the raw input image level. This design makes the learned representations particularly effective for downstream tasks involving small and dense objects. 

\item We propose a stable approach to fuse knowledge from upstream pretraining and domain-specific knowledge learned through unannotated task specific data thorough contrastive learning.
\end{enumerate}

\begin{figure*}[t]
  \centering
   \includegraphics[width=0.9\linewidth]{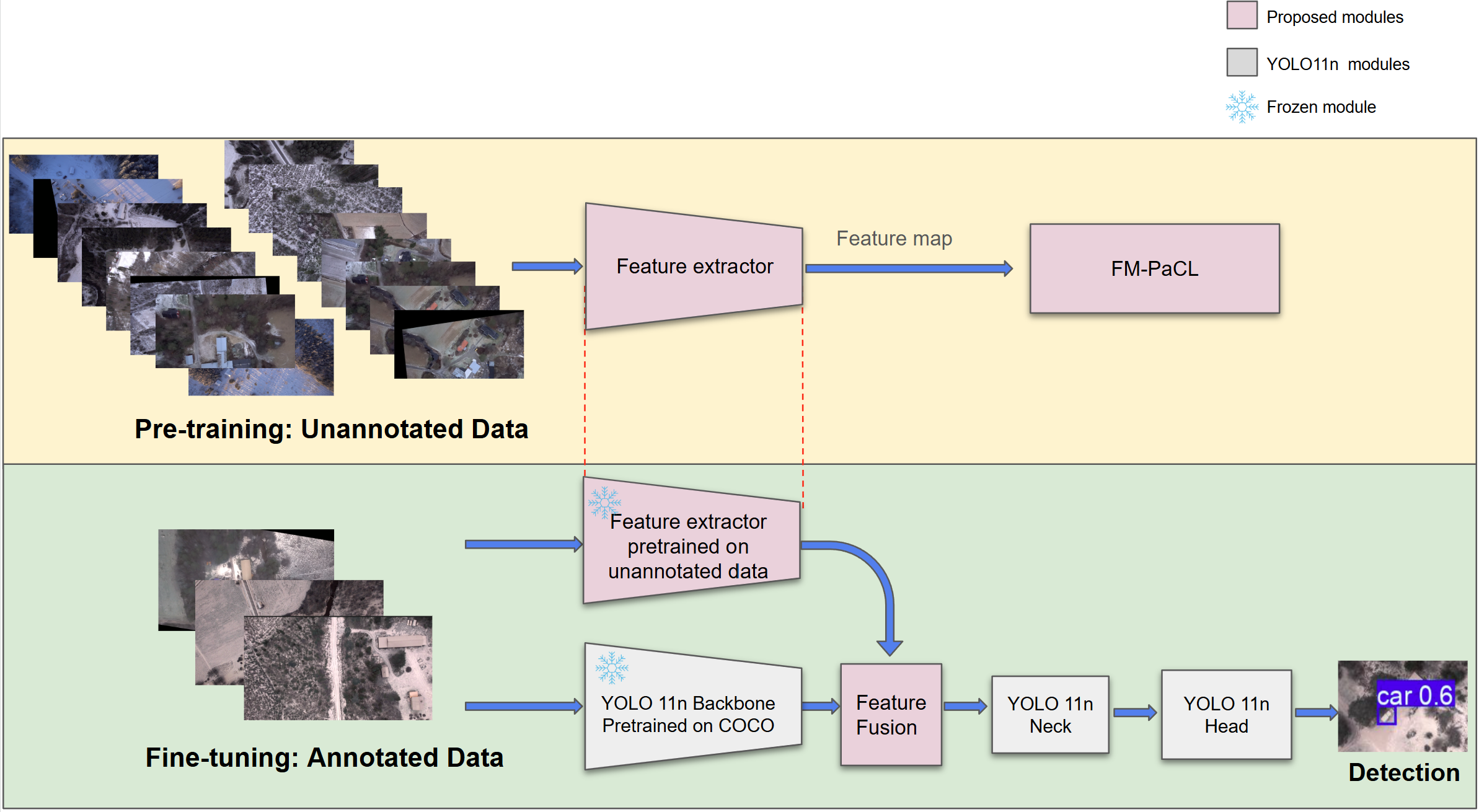} 
   \caption{The proposed Sideload-Contrastive-Learning-Adaption (SCLA) framework consists of two stages: (1) in the pretraining stage, unlabeled data are used to train a CNN-based feature extractor. The intermediate feature maps from this encoder are passed to FM-PaCL (Feature Map Patch-level Contrastive Learning), which enforces the extractor to learn fine-grained, spatially consistent representations by contrasting overlapping patches across augmented views.(2) In the fine-tuning stage, the annotated data feed into both the unannotated data pretrained feature extractor and YOLO11n model. Features from the pretrained feature extractor and YOLO11n's backbone are fused together with a feature fusion block, then the fused features are passed on to YOLO11n's neck and head to output the final detection. During the fine-tuning stage, the pretrained feature extractor and COCO pretrained YOLO11n's backbone are kept frozen. }
   \label{fig:overview}
\end{figure*}

%% file: relatedwork.tex
\section{Related work}

\subsection{Vehicle Detection under Adverse Weather}
Vehicle detection focuses on recognizing and locating vehicles within images or video streams, contributing to improved road safety and more efficient traffic management. Contemporary approaches often rely on deep learning models such as YOLO~\cite{jiang2022review}, Faster R-CNN~\cite{ren2016faster}, and SSD~\cite{liu2016ssd}, which are known for their strong performance in both accuracy and speed. When vehicle detection is applied to data captured by UAVs, it introduces benefits and challenges. UAVs provide a flexible, overhead view that complements traditional ground-based systems. However, aerial perspectives come with difficulties like varying object scales, reduced visual detail, and limited diversity in existing aerial datasets~\cite{li2024toward}. These issues are further amplified under snowy winter weather conditions, where visibility is reduced and image quality may degrade.

The Nordic Vehicle Dataset (NVD)~\cite{nvd} was developed to facilitate research on vehicle detection in challenging Nordic weather conditions. It features a variety of scenes captured from multiple altitudes, including different degrees of snow accumulation and varied cloud cover scenarios. Researchers have tested a broad spectrum of advanced vehicle detection models on this dataset, including single-stage and two-stage detectors, transformer-based approaches~\cite{nvddetr}, and architectures that incorporate wavelet transforms with U-Net~\cite{mokayed2024fractional}. These studies highlight the challenges these deep learning models face when operating in the Nordic environment, particularly when addressing domain shifts caused by diverse weather conditions and variable terrain.

\subsection{Self-Supervised and Contrastive Learning}
In contrast to supervised learning, which depends extensively on annotated datasets to guide model training, self-supervised learning (SSL) derives supervisory signals directly from the data, eliminating the need for external labels. Representation learning strategies are typically categorized into two primary categories: generative and discriminative approaches~\cite{le2020contrastive}. Generative methods aim to reconstruct or model the input data at the pixel level, which is often computationally intensive and not always essential for effective representation learning~\cite{chen2020simple}. Therefore, this work adopts a discriminative strategy to pretrain a feature extractor, specifically contrastive learning, which focuses on distinguishing between similar and dissimilar data instances. The fundamental principle of contrastive learning is to minimize the distance between representations of similar samples while maximizing it for dissimilar ones. Notable frameworks such as SimCLR~\cite{chen2020simple} and MoCo~\cite{he2020momentum} have demonstrated strong performance on large-scale benchmarks like ImageNet~\cite{deng2009imagenet}. Despite its success in learning general-purpose visual features, contrastive SSL encounters limitations when applied to tasks involving small object detection, such as those found in UAV imagery~\cite{kumar2022contrastive}. 

Our proposed FM-PaCL performs contrastive learning at the feature-map patch level rather than compressing the entire input image into a single representation for contrastive learning. We argue that relying on a global image-level embedding is too coarse, preventing the feature extractor from effectively capturing fine-grained local information that is critical for small-object representation. DenseCL~\cite{wang2021dense} introduced dense pixel-level contrastive learning on CNN feature maps can enhance spatial sensitivity. However, DenseCL requires explicit geometric alignment across augmented views, which can be computationally heavy and sensitive to augmentation choices. In contrast, our FM-PaCL extracts overlapping patches from YOLO feature maps and restricts augmentations to photometric ones, ensuring that patch indices are inherently aligned across views. This design not only simplifies training but also focuses the representation learning on fine-grained local cues essential for small-object detection.
\subsection{Feature Fusion Techniques}
In the context of machine learning, feature fusion refers to the integration of features extracted from various layers or sources to construct richer and more discriminative representations. This strategy is particularly beneficial in tasks such as object detection, where combining complementary information from multiple feature hierarchies can significantly enhance model performance~\cite{mungoli2023adaptive, mokayed2021anomaly}. Common fusion techniques include element-wise addition, concatenation, weighted fusion, and attention-based mechanisms~\cite{deng2020feature, dai2021attentional}. While static methods like addition and concatenation treat all features uniformly, dynamic approaches such as attention mechanisms and weighted fusion offer greater adaptability by enabling the model to emphasize more informative features. In the proposed framework, features extracted from a side CNN and those from YOLO's backbone are fused before being passed to the neck and head components of the YOLO architecture. Both static and dynamic fusion strategies have been explored and empirically evaluated. 

\subsection{Transfer Learning and Freezing Strategies}

Transfer learning enables models to utilize knowledge learned from a source domain, often resulting in superior performance compared to training a model from scratch with limited data. One of the most direct approaches to leverage prior knowledge is to reuse the parameters of a pretrained model, as these parameters reflect the model’s learned knowledge~\cite{zhuang2020comprehensive}. To retain this learned knowledge particularly from large-scale datasets, certain layers of the model can be frozen,  Common strategies for freezing include: (i) full freezing, where all pretrained layers remain unchanged and only the final layers are retrained for the new task; (ii) partial freezing, which keeps the early layers (responsible for capturing general features) fixed, while allowing later layers to adapt to the new task; (iii) gradual unfreezing, where layers are progressively unfrozen during training to mitigate catastrophic forgetting~\cite{howard2018universal}; and (iv) full fine-tuning, where the entire pretrained model is retrained on the target dataset. In our proposed framework, both the unannotated NVD data pretrained side CNN and the COCO pretrained YOLO's backbone blocks preceding the fusion stage are frozen during fine-tuning to prevent the pretrained feature extractors from experiencing catastrophic forgetting.


%% file: method_v2_chang.tex
\section{Method}
A pictorial representation of the 
proposed \SideloadFrameworkFull{} (SCLA) framework 
is depicted in Figure \ref{fig:overview}. YOLO11n has been selected as the base detector due to 
an overall consideration of 
accuracy, speed, and resource constraints on edge devices
like UAVs. Our approach begins by pretraining a side CNN using contrastive learning on unannotated data. This pretrained side CNN is then integrated with the YOLO11n model, where its extracted features are adaptively fused with those from the YOLO backbone. The fused features are passed to the neck and head of the detector. To retain the knowledge learned from the unannotated NVD data, the NVD-pretrained side CNN is frozen during the fine-tuning stage. COCO pretrained YOLO11n's backbone also remains frozen to preserve the general visual representations learned from the COCO dataset with a significantly larger scale. 

\input{pretraining}

\input{feature_fusion}

\input{finetune}

%% file: pretraining.tex
\subsection{Pretraining with FM-PaCL}

\begin{figure*}[t]
  \centering
   \includegraphics[width=1.0\linewidth]{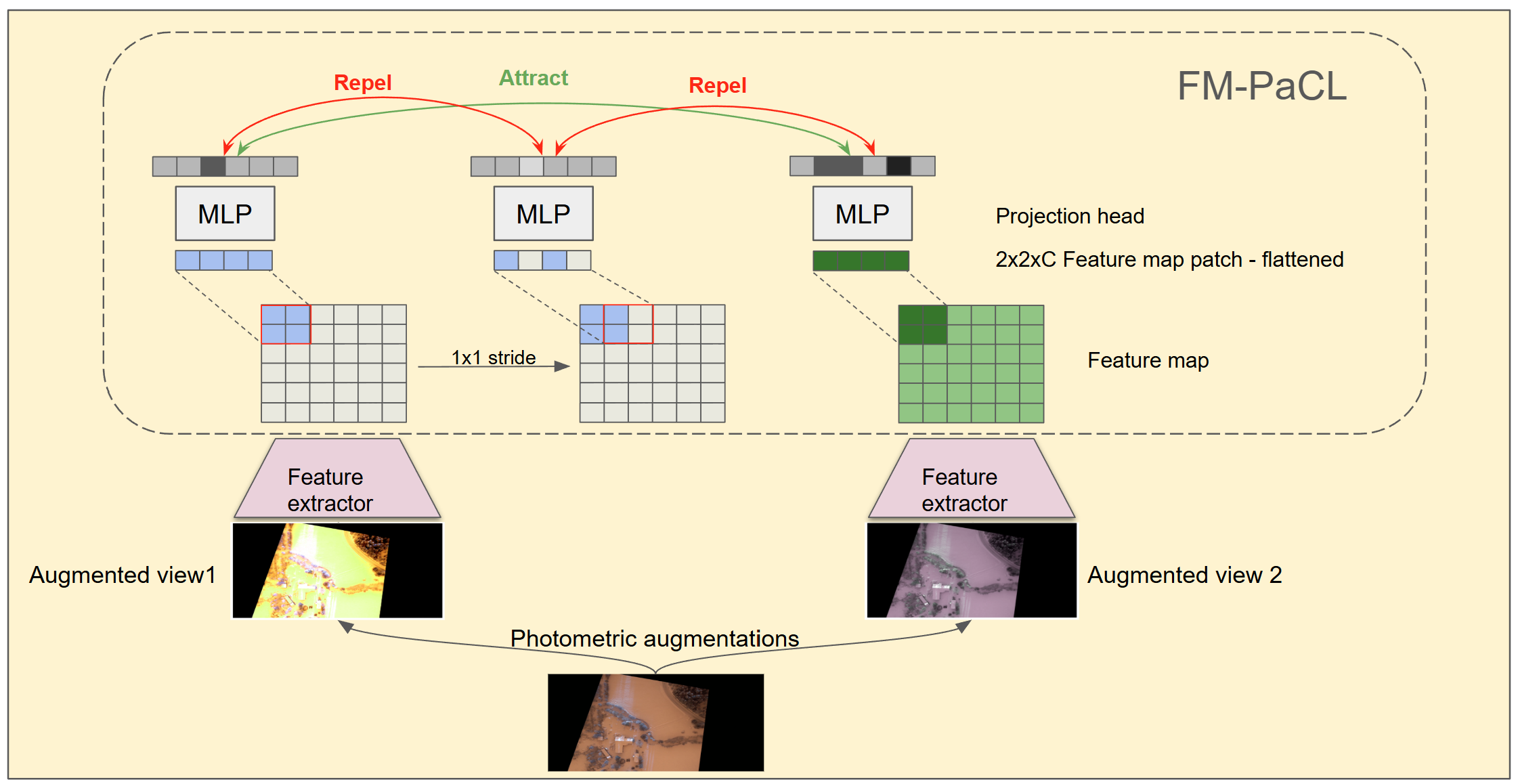} 
   \caption{Illustration of the Pretraining stage. In this stage, only photometric augmentations are applied to preserve the spatial alignment of the feature-map patches. The feature extractor is a CNN with the same architecture as YOLO11n's backbone, ensuring that the resulting feature-map dimensions match those of YOLO11n and enabling seamless integration during the Feature Fusion stage. C denotes the number of channels in the feature map.}
   \label{fig:FM-PaCL}
\end{figure*}

We propose \textbf{FM-PaCL} (Feature Map Patch-level Contrastive Learning), a self-supervised pretraining method that performs contrastive learning on overlapping patches extracted from intermediate feature maps of a side CNN. Unlike prior contrastive frameworks that compress the entire image into a single global embedding, FM-PaCL preserves spatial correspondence and local detail by operating directly on the encoder’s high-resolution feature grid (see Fig.~\ref{fig:FM-PaCL}).

\subsubsection{Patch Extraction from Feature Maps}
Given an input image $x \in \mathbb{R}^{3 \times H \times W}$, we pass it through the side CNN to obtain a feature map $f_x \in \mathbb{R}^{C \times H_f \times W_f}$. 
To construct local feature map patches for contrastive learning, we apply a sliding window of size $(p_h, p_w)$ with stride $(s_h, s_w)$ to partition the feature map into overlapping patches. Each patch is flattened and projected through a small MLP head $g(\cdot)$, producing patch embeddings $\mathbf{z}_{i,j} \in \mathbb{R}^d$.
The result is a set of $P = H_p \times W_p$ embeddings per image, where $H_p$ and $W_p$ denote the number of patch positions.

\subsubsection{Contrastive Loss}
We adopt a Siamese setup with two augmented views of each image. Augmentations are restricted to \textit{photometric transformations} (e.g., color jitter, Gaussian blur, grayscale) to ensure that the spatial alignment of patches across views remains intact. For a batch of size $B$, each image contributes $P$ patch embeddings per view, resulting in $2BP$ embeddings in total. We employ a patch-level InfoNCE loss. Each patch has one positive match, which is the patch at the same patch position from the other view of the same image with different photometric transformations. While all other patches from the mini-batch act as negatives. This formulation enforces invariance to photometric variations while preserving positional correspondence.

Similarity between two patch embeddings is computed by temperature-scaled dot product and denoted as sim$(u, v)$. For a positive patch pair, i.e., the same image and the same patch 
index across the two augmented views, we denote the corresponding patch 
embeddings as $z^{(1)}_{i,p}$ and $z^{(2)}_{i,p}$, where $i$ indexes the 
underlying image and $p \in \{1,\dots,P\}$ indexes the patch location. 
The InfoNCE loss for the positive pair of ($z^{(1)}_{i,p}$, $z^{(2)}_{i,p}$) is defined as

\[
\ell_{(i,p)}^{1 \rightarrow 2}
=
-
\log
\frac{
\exp\!\left( 
\operatorname{sim}\!\left( z^{(1)}_{i,p},\, z^{(2)}_{i,p} \right)
/ \tau
\right)
}{
\sum\limits_{j=1}^{B}
\sum\limits_{q=1}^{P}
\mathds{1}_{\{(j,q)\neq(i,p)\}}\,
\exp\!\left( 
\operatorname{sim}\!\left( z^{(1)}_{i,p},\, z^{(2)}_{j,q} \right)
/ \tau
\right)
}.
\]

The final loss is computed across all positive pairs, both ($z^{(1)}_{i,p}$, $z^{(2)}_{i,p}$)
and ($z^{(2)}_{i,p}$, $z^{(1)}_{i,p}$), in a mini-batch. The symmetric patch-level contrastive loss of the mini-batch is then
\[
\mathcal{L}
=
\frac{1}{2N'}
\sum_{i \in \mathcal{I}}
\sum_{p=1}^{P}
\left(
\ell_{(i,p)}^{1 \rightarrow 2}
+
\ell_{(i,p)}^{2 \rightarrow 1}
\right),
\]
where $\mathcal{I}$ is the set of positive-image indices and 
$N' = |\mathcal{I}|\,P$ is the total number of patch positions drawn from positive images.

The side CNN and the MLP head $g(\cdot)$ are updated to minimize $\mathcal{L}$. After the contrastive pretraining, $g(\cdot)$ is thrown away, and the unannotated NVD pretrained side CNN will be side-loaded in the fine-tuning stage.

%% file: feature_fusion.tex
\input{pic/train_dataset}

\subsection{Feature Fusion}
In the finetuning stage, the following feature fusion methods have been implemented to integrate features from the side CNN, which is pretrained on unannotated data, with those extracted by the COCO-pretrained YOLO11n backbone. 
The side CNN is implemented with the same architecture as the YOLO11n's backbone, the features from the side CNN and the features from the YOLO backbone therefore have the same spatial size ($H$x$W$) and channel number($C$), which enables a seamless fusion. 
\begin{figure*}[!ht]
  \centering
  \begin{subfigure}{0.3\linewidth}
    \includegraphics[width=\columnwidth]{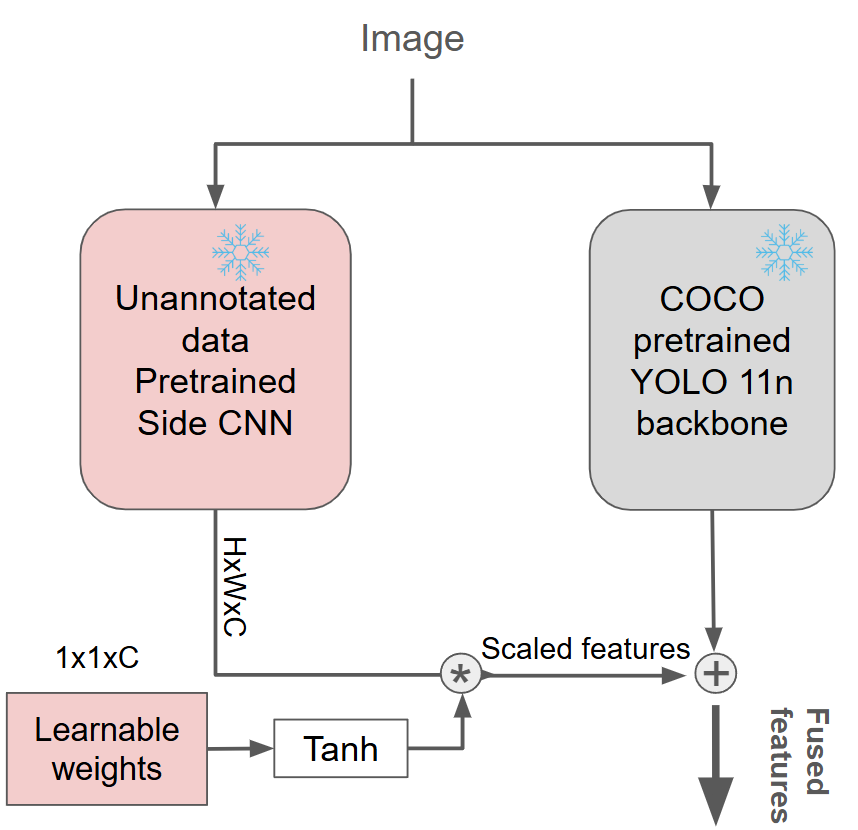}
    \caption{Learnable weights gating}
    \label{fig:learnable_w}
  \end{subfigure}
  \hfill
    \begin{subfigure}{0.3\linewidth}
    \includegraphics[width=\columnwidth]{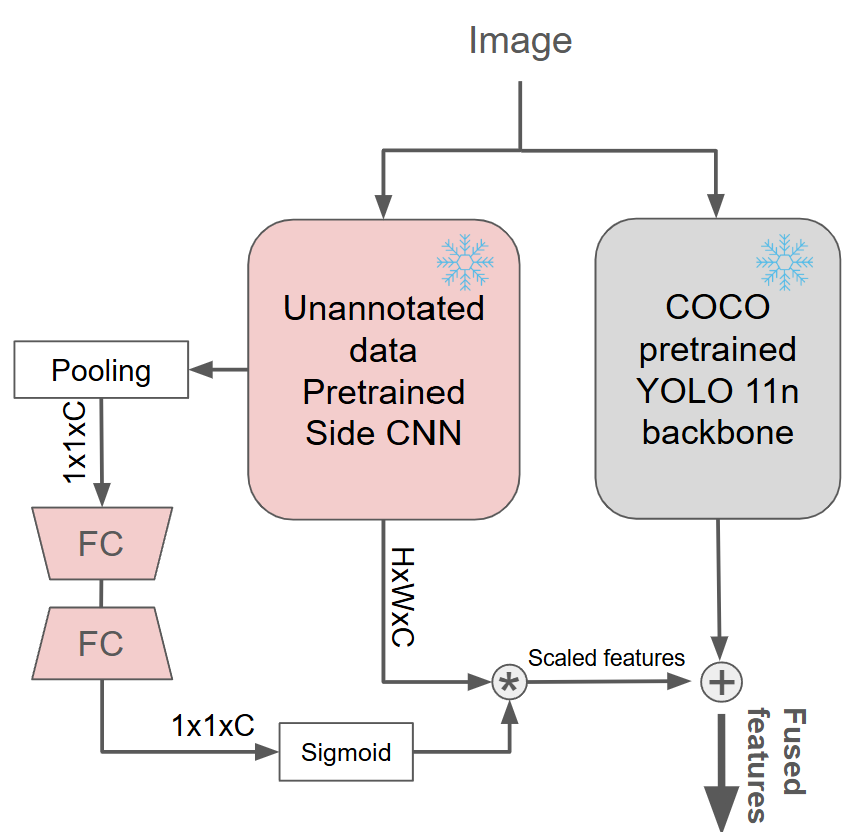}
    \caption{SE gating}
    \label{fig:SE}
  \end{subfigure}
  \hfill
  \begin{subfigure}{0.3\linewidth}
    \includegraphics[width=\columnwidth]{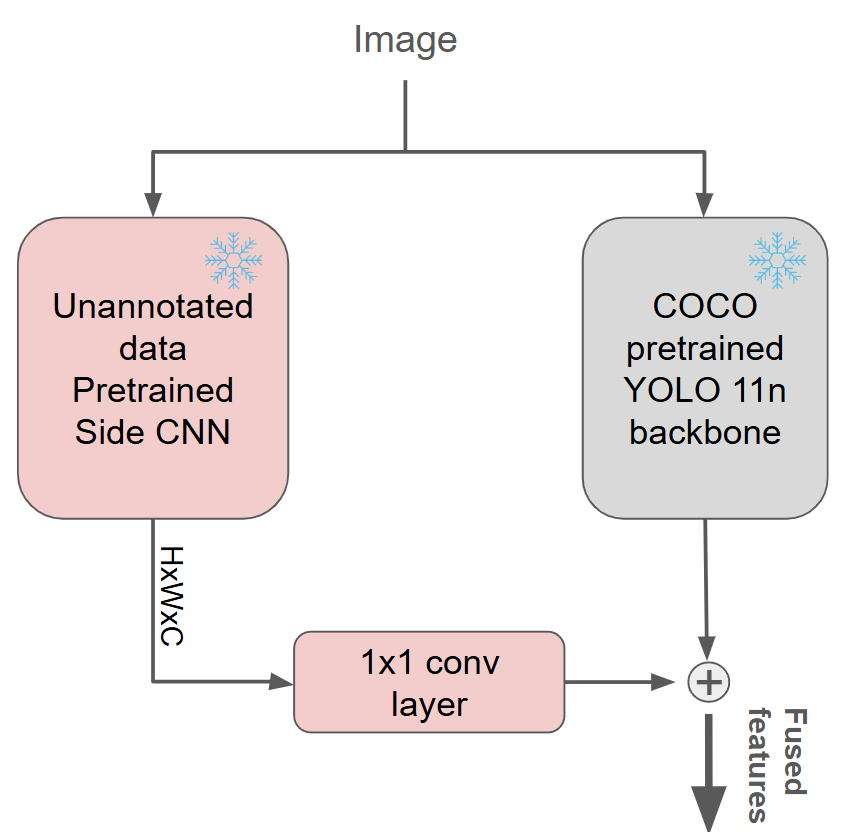}
    \caption{Zero-conv gating}
    \label{fig:zeroconv}
  \end{subfigure}
  \caption{Comparison of different gating mechanisms for feature fusion, including learnable weights, SE blocks, and Zero-Conv layers. \newline
  \footnotesize * means channel-wise multiplication between the scalar and the feature map of the side CNN. + represents element-wise adding.}
  \label{fig:fusionblock}
\end{figure*}

In this work, we compare the following fusion schemes: 

\subsubsection{Addition}
The addition method involves summing corresponding elements of feature maps. It requires the feature maps to have the same dimensions. 


\subsubsection{Learnable weights gating}
This method involves taking a weighted sum of feature maps, where weights are learned during training. The weights dimension is 1x1x$C$, where $C$ is the number of channels. This allows the model to learn the importance of feature channel-wise. The learnable weights $W$ are initialized to 0, and squashed by the tanh activation function. In the first training step, $tanh(W)$ will output 0. This means that the fused feature maps, at step 0, are identical to those extracted from the upstream YOLO11n backbone pretrainied on COCO. This initiation enables the framework to gradually adapt to features learned from domain-specific images via FM-PaCL on unannotated data.

\subsubsection{SE gating}
This method adopts a Squeeze and Excitation~(SE) block \cite{hu2018squeeze} to gate the features from the side CNN. SE is a channel-wise self-attention mechanism.
This mechanism allows the network to weigh the per-sample importance of different channels in the feature maps produced by the Side CNN and upstream COCO pretraining.
The squeeze operation reduces the spatial dimensions of the feature maps($H$x$W$x$C$) to a single value per channel(1x1x$C$). Then, the excitation operation  
uses these squeezed values, i.e., the $1\times1\times C$ channel descriptor, to generate weights for each channel. These weights are then multiplied by the original feature maps to generate the scaled feature map, which emphasizes important channels and suppresses less important channels. To maintain the efficiency of the model, the gating mechanism is implemented by a bottleneck architecture containing two fully-connected (FC) layers with a ReLU activation function in between. As such, with reduction ratio $r$, the first FC reduces the dimensionality by projecting the descriptor from 1x1x$C$ to 1x1x($C/r$), followed by a ReLU and then the second FC increases the dimension back to 1x1x$C$~\cite{hu2018squeeze}. In this work, the reduction ratio $r$ is set as 16 following Hu et al.~\cite{hu2018squeeze}.

\subsubsection{Zero-Conv}
Inspired by Control Net~\cite{zhang2023adding}, we also adopt zero-conv layers to fuse features from the side CNN pretrained via FM-PaCL and the upstream backbone pretrained on the COCO dataset. 
Specifically, the Zero-Conv layer is a 1×1 convolution layer. The initiation of Zero-Conv is similar to Learnable weights gating, with both weight and bias initialized to zeros,
so that the fused feature maps start identical to those extracted with the COCO-pretrained backbone, and progressively adapt features to the FM-PaCL pretrained side CNN.



\input{pic/bench}

%% file: pic/train_dataset.tex
\begin{table*}[t]
    \centering
    \footnotesize
    \caption{Information on the videos and their roles  (training, validation and testing) in ours and Mokayed et al.'s protocol (NVD)~\cite{nvd}. }
    \begin{tabular}{l|l|l|l|l|l|l|l|l}
    \hline
        \textbf{Video}&\textbf{Frames}&\textbf{Split-ours}&\textbf{Split-NVD~\cite{nvd}}& \textbf{Altitude} & \textbf{Snow cover} & \textbf{Cloud cover} & \textbf{fps} & \textbf{GSD} \\ \hline
        Asjo 01 &801 &Train &641 for Train, 160 for Val& 130–200 m & minimal (0–1 cm) & overcast & 5 & 11.5–17.8 cm \\ 
        Bjenberg &4003 &Train&3203 for Train, 800 for Val& 250 m & Fresh (1–2 cm) & light & 25 &22.2 cm \\
        Asjo 01 HD &1252&Train&1002 for Train, 250 for Val& 250 m & Fresh (5–10 cm) & clear & 5 & 20.2 cm\\ \hline
        Nyland 01 &1203&Val&Test& 150 m & Minimal (0–1 cm) & Dense & 5 &11.5–17.8 cm \\ \hline
        Bjenberg 02&1191&Test&Test&250 m &Fresh (5–10 cm) &clear& 5 &11.1cm \\ \hline

    \end{tabular}
    \label{tab:trainsplit}
\end{table*}

%% file: pic/bench.tex
\begin{table*}[t]
    \centering
    \caption{NVD dataset~\cite{nvd} performance with Mokayed et al.'s  protocol (NVD)~\cite{nvd} and our protocol. Yolo11n is full parameter fine-tuning Yolo11n on the NVD dataset.}
    \begin{tabular}{l|l|c|c|c|c|c}
    \hline
    \textbf{Name} & \textbf{Protocol} & \textbf{Precision} & \textbf{Recall} & \textbf{F-measure} & \textbf{mAP@50} & \textbf{Std of mAP@50} \\
    \hline
    DETR-Refined~\cite{nvddetr} & NVD~\cite{nvd} & \textbf{85.4} & \textbf{70.2} & \textbf{77.1} & \textbf{79.4} & - \\
    \hline
    Yolov8s-Au~\cite{nvd} & NVD~\cite{nvd} & \textit{77.1} & 34.6 & 47.8 & 50.7 & - \\
    Yolo11n   & NVD~\cite{nvd} & 74.1 & 49.2 & 58.6 & 61.3 & 2.9 \\
    Ours & NVD~\cite{nvd} & 75.5 & \textit{63.4} & \textit{68.9} & \textit{70.2} & 2.6 \\
    \hline
    \hline
    Yolo11n & Ours & 54.0 & 45.0 & 48.3 & 46.6 &  2.0\\
    Ours & Ours & 65.4 & 56.0 &60.3  & 58.2 & 2.1 \\
    \hline
    \end{tabular}
    \label{tab:bench}
\end{table*}
\begin{table*}[t]
    \centering
    \caption{Ablative studies on the framework level. We report the average performance with 3 different runs and the standard deviation of the mAP@50 metric. Yolo11n is full parameter fine-tuning Yolo11n on the NVD dataset, PaCL pretrained backbone is partial fine-tuning
    Yolo11n with Yolo11n’s backbone pretrained on unannotated NVD frozen, Frozen Backbone is partial fine-tuning Yolo11n with Yolo11n's backbone frozen. Unannotated NVD indicates the model utilizes the knowledge from unannotated data through contrastive learning, COCO indicates the model inherent upstream feature extractor pretrained on coco via backbone freezing.  }
    \begin{tabular}{l|c|c|c|c|c|c|c}
    \hline
    \textbf{Name} &\textbf{Unannotated NVD}&\textbf{COCO}& \textbf{Precision} & \textbf{Recall} & \textbf{F-measure} & \textbf{mAP@50} & \textbf{Std of mAP@50} \\
    \hline
    Yolo11n & & &54.0 & 45.0 & 48.3 & 46.6 & 2.0 \\
    PaCL pretrained backbone  &\checkmark& & 58.2 & 40.7 & 47.9 & 42.4 & 1.7 \\
    Frozen Backbone&&\checkmark& 63.9 & 53.3 & 58.0 & 55.4  & 3.8 \\
    \hline
    Ours &\checkmark&\checkmark& \textbf{65.4} & \textbf{56.0} & \textbf{60.3} & \textbf{58.2} & 2.1 \\
    \hline
    \end{tabular}
    \label{tab:ablativeframe}
\end{table*}


%% file: finetune.tex
\subsection{Fine-tuning}
In the fine-tuning stage, only annotated data is used to supervise the training of the SCLA framework. The feature extractors, namely the sideload feature extractor pretrained via FM-PaCL on unannotated data, and the YOLO11n backbone pretrained on COCO, are kept frozen to preserve the representations learned from pretraining. The augmentation pipeline applied in the fine-tuning stage consists primarily of photometric transformations and mild geometric variations such as translations, scaling, and random horizontal flipping. To further improve robustness, we include RandAugment as the overall augmentation policy, enable MixUp for sample-level regularization, and apply random erasing as an additional form of local perturbation.

%% file: experiments_251223.tex
\section{Experiments}

\subsection{Implementation details}

Different from Mokayed et. al's Protocol (NVD)~\cite{nvd}, which splits the same videos into train and validation\footnote{the testing frames in Mokayed et. al's Protocol are from different videos used in train and val.}, we split the annotated data in the NVD dataset into train, validation, and test sets, on \textbf{individual video level}, and use it as our main protocol. Specifically, the training dataset includes 3 videos (Asjo 01, Bjenberg and Asjo 01 HD) with 6056 annotated frames. The validation set includes 1 video (Nyland 01) with 1203 annotated frames. The testing set includes 1 video (Bjenberg 02) with 1191 annotated frames, recorded on a different date and at a different location from those used in the training and validation sets. 
This setup introduces a domain gap, providing a more realistic assessment of the model's generalization capability.  
The exact split and video information are shown in Table~\ref{tab:trainsplit}.

In the fine-tuning stage, we select the model at the iteration where it achieves the best validation performance, and use it for the evaluation on the test set.
We also benchmark our proposed method on the original split for fair comparisons against existing methods.

\subsection{Benchmark}

In this section, the proposed \SideloadFrameworkFull{} framework with SE gating is compared to YOLO detection methods~\cite{nvd,nvddetr} on their protocol, and then we discuss the differences between their protocol and our proposed protocol. The exact test split is recovered from the weights and logs released in~\cite{nvd}. For training and validation, we split the first 80\% frames of each non-testing video as the training set and the last 20\% of each non-testing video as the validation set. The average results of 3 independent runs with standard deviation are shown in Table~\ref{tab:bench}. 

The results of the original NVD protocol~\cite{nvd} indicate that our proposed framework improves the mAP50 performance of YOLO11n by 8.9\% (Yolo11n vs Ours), with more consistent performance over multiple runs.
The proposed method (Ours) is 19.5\% better than existing CNN-based methods so far~\cite{nvd} on mAP50 (Yolov8s-Au vs Ours), which makes ours promising to be deployed on edge devices with strong resource constraints.  
Although performance-wise weaker than the DETR-based detector~\cite{nvddetr}, the smaller size of the YOLO11n model makes our method a better fit to be deployed on the UAVs with limited computation power and/or internet connection.      

Compared to the NVD~\cite{nvd} protocol, our split protocol offers a more realistic evaluation setting, particularly in scenarios involving significant domain shifts. Under our protocol, the baseline YOLO11n model experiences a performance drop of 14.7\% in mAP50 when switching from the NVD protocol. While our proposed framework maintains a comparable level of performance, highlighting its robustness to domain variation.

\subsection{Ablative study}
In this section, we discuss the following research questions:
\begin{itemize}
    \item[] \textbf{RQ1}: \textit{What training/pretraining sources are important to the NVD task?}
    \item[] \textbf{RQ2}: \textit{If multiple sources of pretrained features are beneficial, how to fuse them effectively?} 
    \item[] \textbf{RQ3}: \textit{What level of fusion granularity do we need?} 

\end{itemize}

\subsubsection{What training/pretraining sources are important to the NVD task?}
This question is broken down into the following three subquestions: 
\begin{itemize}
    \item[] \textbf{RQ1a}: Is COCO pretraining needed? 
    \item[] \textbf{RQ1b}: Is pretraining via PaCL on unannotated data needed? 
    \item[] \textbf{RQ1c}: Is pretraining via PaCL on unannotated data all you need? 
\end{itemize}

\paragraph{RQ1a: Is COCO pretraining needed?} 
We answer this through comparing the baseline Yolo11n, where all parameters are fully fine-tuned on the annotated NVD dataset, with the Frozen Backbone setting, in which the YOLO11n backbone pretrained on COCO is kept frozen during fine-tuning to preserve upstream COCO representations. Retaining the COCO-pretrained backbone yields an 8.8\% (Table~\ref{tab:ablativeframe}) improvement in mAP50 compared to fully fine-tuning Yolo11n, demonstrating the importance of COCO pretraining.

\paragraph{RQ1b: Is pretraining via PaCL on unannotated data needed?} 
We answer this through comparing the upstream COCO feature only ``Frozen Backbone'' and the proposed framework ``Ours'' with additional features from the PaCL pretraining on unannotated data. ``Ours'' outperforms ``Frozen Backbone'' by 2.8\% in mAP50.

\paragraph{RQ1c: Is pretraining via PaCL on unannotated data all you need? } 
We answer this through using ``PaCL  pretrained backbone'' setting. In this setup, the backbone network is first pretrained on unannotated data using FM-PaCL. Afterward, the pretrained backbone is frozen, the original neck and head are reattached, and only the neck and head are fine-tuned on annotated data. However, ``PaCL pretrained backbone'' does not achieve satisfactory performance. This under-performance may be due to catastrophic forgetting of the COCO-pretrained features, and the patch-level contrastive pretraining emphasizing local feature discrepancies. As a result, features learned solely from unannotated NVD data are not well aligned with the requirements of downstream object detection tasks.

\paragraph{Answer to RQ1: What training and pretraining sources are important for the NVD task?}
Taken together, the results from \textit{RQ1a--RQ1c} demonstrate that both upstream COCO pretraining and PaCL pretraining on unannotated data are essential for strong NVD vehicle detection performance. While COCO pretraining provides robust and transferable visual representations, PaCL pretraining further enriches the model with domain-specific features learned from unannotated data. Combining these two sources yields an 8--10\% improvement in mAP50 on both evaluation protocols.

\begin{figure}
\centering
\includegraphics[width=\columnwidth]{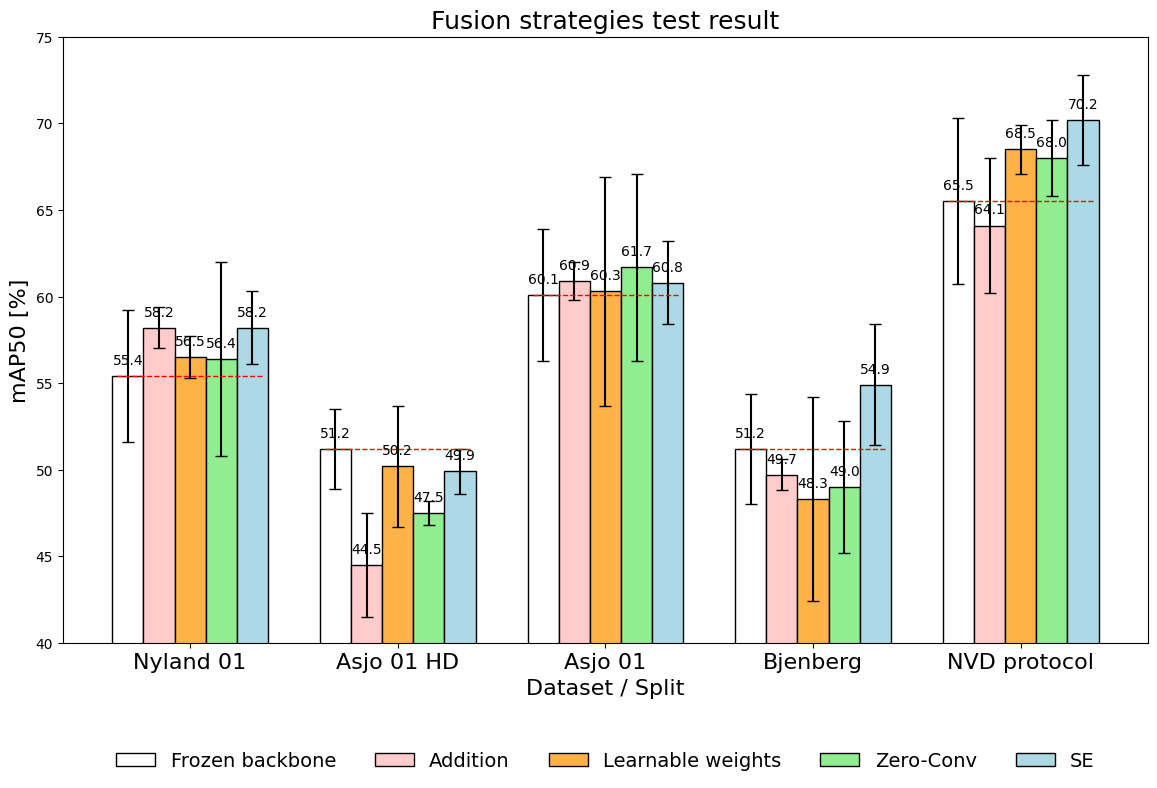}
\caption{\textcolor{black}{Test results of different fusion techniques. Each group of bars represents the fusion in the order of 'Frozen backbone', 'Addition', 'Learnable weights', 'Zero-Conv','SE'. Better viewed in color and zoomed in.}}
\label{fig:results}
\end{figure}

\subsubsection{How to fuse multiple sources features effectively?} 

In the sideload approach, a key challenge is how to effectively integrate features from the side CNN pretrained via FM-PaCL on unannotated data with those from YOLO's backbone pretrained on COCO. To address this, we explored and evaluated the following fusion strategies: "Addition", "Learnable Weights gating", "SE gating", and "Zero-Conv gating". Given the limited amount of annotated data, only five videos, we conduct cross-validation to evaluate the fusion strategies across multiple data splits. This approach provides a more reliable estimate of performance and reveals how consistent different fusion techniques are across different subsets. We consistently use the video ’Bjenberg 02’ as the test set, while the
remaining four videos are used in a leave-one-out fashion for validation. Together with Mokayed et. al’s split Protocol (NVD), we conduct experiments on five different data splits. For each split, we only pretrain the side CNN using unannotated videos\footnote{certain unannotated videos are very similar to some annotated ones, so we excluded them manually for each split to prevent data leakage} not associated with validation and testing to ensure no validation or testing data is used during pretraining.
Each configuration runs with 3 different seeds, the average mAP50 and the standard deviation of 3 runs are shown in Figure~\ref{fig:results}. Overall, SE gating outperforms “Frozen backbone” baseline by an average of 2.2\% across the five splits. Notably, SE gating surpasses the “Frozen backbone” by 3.7\% and 4.7\% mAP50 on the Bjenberg-validation split and the NVD split, respectively. These two splits contain less training data during fine-tuning, suggesting that when annotated data is scarce, our sideload approach(SCLA) yields greater performance gains. Learnable weights and zero-conv achieve a comparable performance to "Frozen backbone" in terms of mAP50 on average of the 5 splits. Our results demonstrate that the pretrained side CNN, together with SE gated fusion, performs the best within the scope of our experiment, while also exhibiting more consistent results across different seeds compared to the “Frozen backbone”. We credit this advantage to SE gating's capability to dynamically adjust the features' importance channel-wise, w.r.t the input and task objectives. 

\subsubsection{What fusion granularity level is needed?}

\begin{figure}
\centering
\includegraphics[width=0.8\linewidth]{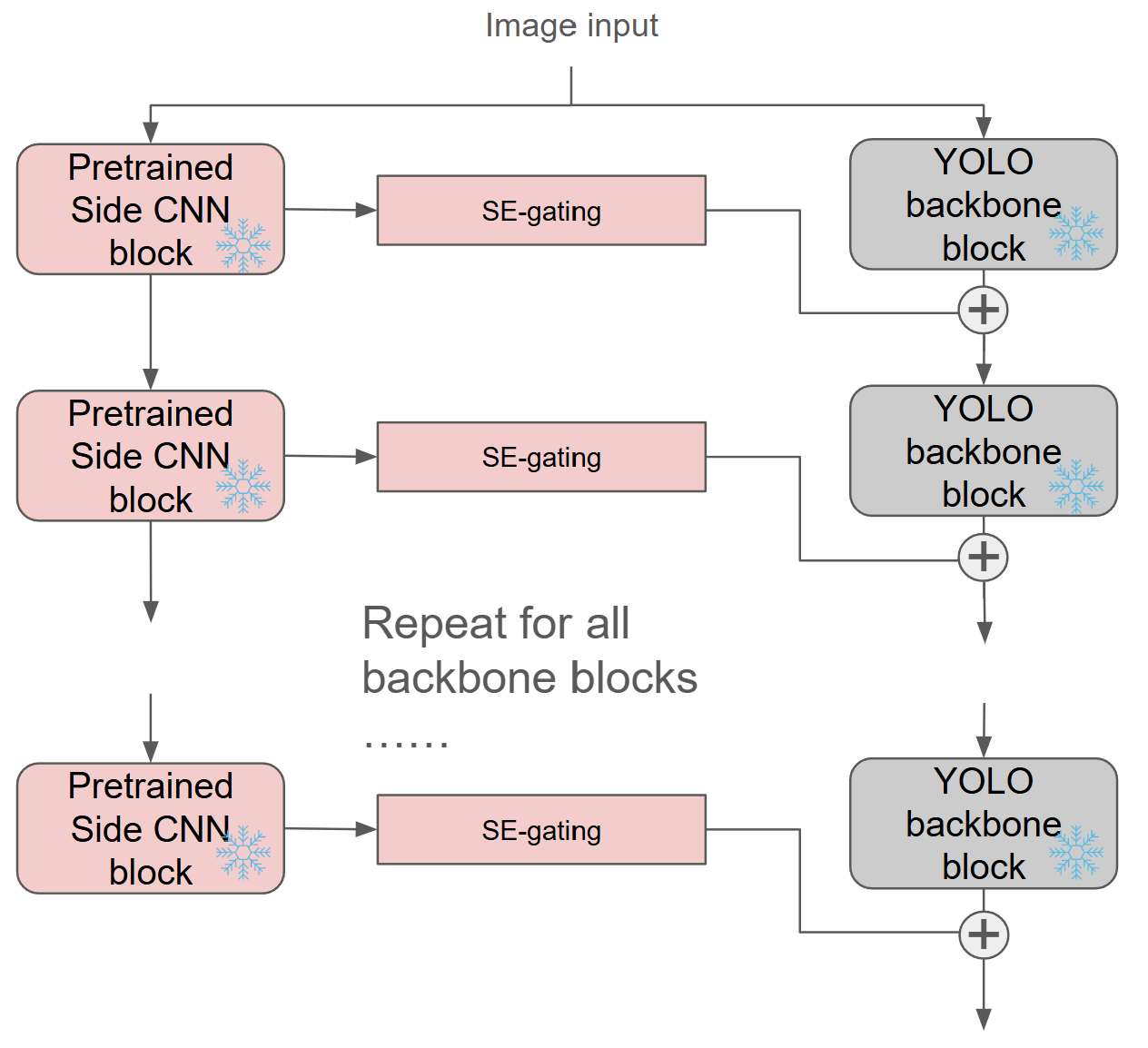}
\caption{\textcolor{black}{Blockwise feature fusion}}
\label{fig:blockwise_model}
\end{figure}

To further investigate the potential performance improvements, blockwise feature fusion is implemented for each fusion technique presented earlier. Rather than fusing features only at the end of the backbone,
blockwise-fusion integrates the features from the side CNN and YOLO backbone at
each block stage, an example of blockwise SE gated fusion is illustrated in Figure~\ref{fig:blockwise_model}. Despite its increased complexity, the blockwise-fusion architecture
does not yield performance improvements. Instead, the test results show a decline compared to "Frozen backbone". Among the fusion strategies, blockwise addition performs the worst, while gating-based fusion methods show improved results over direct addition but still fall short of "Frozen backbone" performance.  This result indicates that combing features too early or too frequently may mix incompatible representations, possibly leading to noisy signals. Additionally, injecting external features at every block can disrupt the YOLO backbone's hierarchical learning, making it harder for the network to learn meaningful representations. Fusing only at the end of the backbone allows the model to combine global context without disrupting earlier feature extraction stages.

%% file: limitation.tex
\section{Limitation}
\begin{figure}
\centering
\begin{subfigure}[t]{0.45\textwidth}
    \centering
    \includegraphics[width=\linewidth]{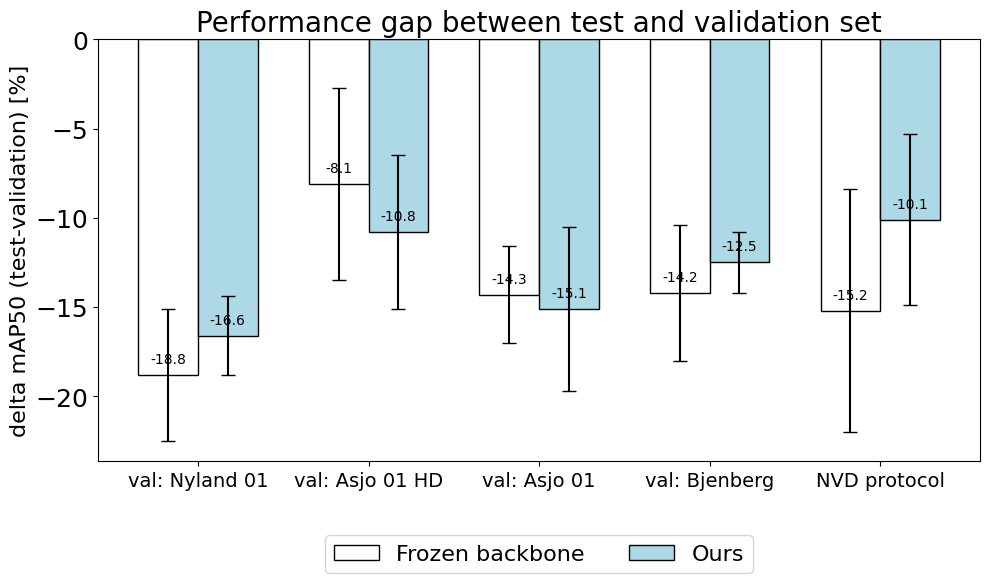}
    \caption{\textcolor{black}{Cross-validation: validation and test result gap.}}
    \label{fig:val-test_gap}
\end{subfigure}
\hfill
\begin{subfigure}[t]{0.45\textwidth}
    \centering
    \includegraphics[width=\linewidth]{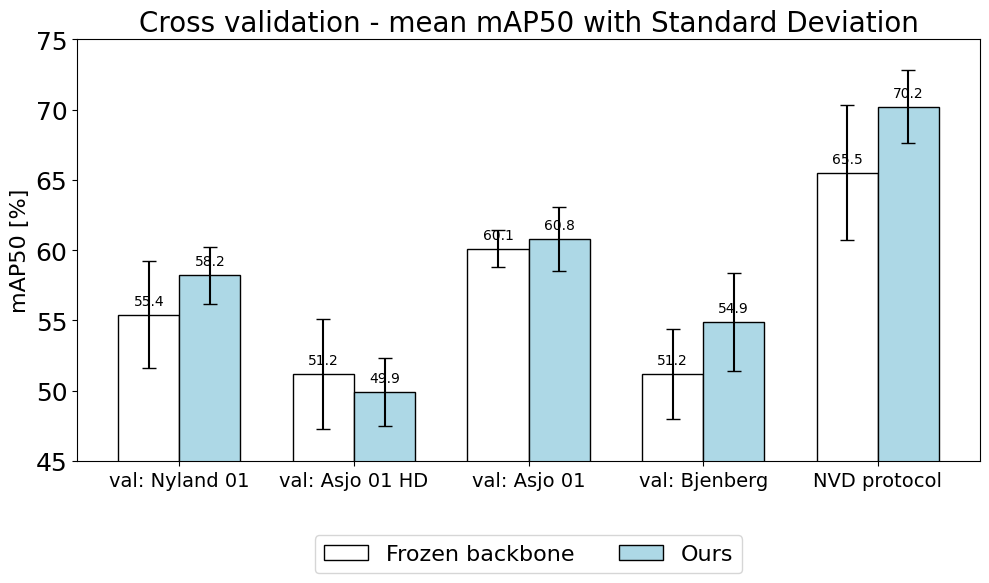}
    \caption{\textcolor{black}{Cross-validation test results.}}
    \label{fig:results_crossval}
\end{subfigure}
\caption{Cross-validation performance analysis. Each label on the horizontal axis indicates the video used as the validation set. For example, 'val: Nyland 01' means 'Nyland 01' was used for validation, 'Asjo 01 HD', 'Asjo 01' and 'Bjenberg' were train set. The test set remains 'Bjenberg 02' for splits of 'val: Nyland 01', 'val: Asjo 01', 'val: Asjo 01 HD' and 'val: Bjenberg'.}
\label{fig:crossval_combined}
\end{figure}

Even though our proposed SCLA framework demonstrates an overall improvement over "Frozen backbone", there is an exception of the split where 'Asjo 01 HD' is used as the validation set. To better understand this anomaly, we checked the video characteristics and the gap between the validation and the test performance. The gap between validation and test performance is visualized in Figure~\ref{fig:val-test_gap}. The observation is that our method particularly improves test performance when "Frozen backbone" exhibits a large gap between validation and test performance. In terms of weather conditions, 'Asjo 01 HD' closely resembles the test set 'Bjenberg 02', featuring fresh snow coverage and clear skies, suggesting that the domain shift between these two splits may be smaller than in other data splits. However, due to the limited amount of annotated data, it is difficult to draw a definitive conclusion about whether this exception is caused by domain differences in the data split.

%% file: conclusion.tex
\section{Conclusion}
Weather factors such as varying cloud coverage and differing levels of snow accumulation introduce significant domain gaps, which pose substantial challenges for UAV-based object detection systems. In this work, we proposed a contrastive learning-enhanced framework that builds upon the lightweight YOLO11n to address these challenges. Our approach mitigates the domain gap under both the NVD protocol~\cite{nvd} and a more practical dataset split protocol without the need for additional manual annotations. Extensive experiments on the proposed Sideload-Contrastive-Learning-Adaption (SCLA) framework, together with detailed result analysis, demonstrate its effectiveness.
In summary: 
\begin{itemize}
    \item \textit{Do we need features from multiple upstream sources?} Yes. We need both upstream COCO pretraining and NVD unannotated data pretraining to achieve strong performance under the NVD setting.
    \item \textit{How to pretrain the feature extractor on NVD unannotated data?} Using FM-PaCL. This approach applies contrastive learning to feature-map patches, making the learned representations particularly effective for downstream tasks involving small and dense objects in the UAV-captured imagery.
    \item \textit{How to fuse the features from different sources?} SE gating, the channel-wise self-attention mechanism, shows the best within the scope of our experiment.
    \item \textit{What fusion granularity level is needed?} Feature fusion is most effective when performed only at the end of the backbone, rather than at every backbone layer. Backbone-level fusion enables the model to integrate global contextual information without disrupting early-stage feature extraction.
\end{itemize}

\section{Future work}

Our findings suggest that blockwise feature fusion appears to negatively impact YOLO detection performance. An open question remains regarding its influence on training loss and whether some bad outsourcing behaviors occur in specific blocks. Key areas for future work include distinguishing between beneficial and detrimental outsourcing behaviors, developing assessment metrics during training, and exploring regularization strategies to mitigate negative effects. These efforts aim to refine training methodologies and support the development of more robust and interpretable representation learning systems. 

Additionally, the exceptional performance drop observed when using 'Asjo 01 HD' video as the validation split is worth further investigation. For example, whether this performance drop is related to any unique characteristics of the 'Asjo 01 HD'. Understanding this behavior could provide a deeper understanding of domain gaps and improve the generalization of detection models under diverse environmental conditions.

\label{Chapter:Conclusion}